%% file: iclr2027_conference.tex
\documentclass{article} % For LaTeX2e
\usepackage{iclr2027_conference,times}

\input{math_commands.tex}

\usepackage{hyperref}
\usepackage{url}
\usepackage{times}
\usepackage{latexsym}
\usepackage{dsfont}
\usepackage[T1]{fontenc}
\usepackage[utf8]{inputenc}

\usepackage{microtype}

\usepackage{inconsolata}

\usepackage{graphicx}
\usepackage{amsmath,amssymb}
\usepackage{booktabs}
\usepackage{placeins}
\usepackage{graphicx}
\usepackage{microtype}
\usepackage{inconsolata}
\usepackage{enumitem}
\usepackage{wrapfig}

\author{Yuanxiang Huangfu \& Hanmeng Zhong \&
Linqing Chen \& Jeffrey Tiong Jee Hui \\
PatSnap Co., LTD.\\
Suzhou, China \\
\texttt{\{huangfuyuanxiang,zhonghanmeng,chenlinqing,jtiong\}@patsnap.com} \\
 }

\title{From Retrieval to Recognition:\\
How Vision--Language Models Become OCR Specialists}
\iclrfinalcopy
\begin{document}
\maketitle
\lhead{Under review as a conference paper at ICLR 2027}  % 清除 “Published as a conference paper at ICLR 2027”
\begin{abstract}
Does a general vision--language model acquire specialized OCR ability by developing a new reading circuit or by reusing an existing mechanism? We address this question in the setting of full-sequence OCR, rather than local-answer retrieval. Using an evidence-grounded protocol with held-out causal interventions, we identify sparse and stable OCR-head sets in GLM-OCR, MinerU2.5, and PaddleOCR-VL-1.6. We then investigate the mechanistic origin of these OCR heads by comparing them with independently identified textual retrieval/copy heads in general VLMs. Across two general VLMs, visual OCR heads strongly overlap independently identified textual retrieval/copy heads, yielding untuned top-20 intersections of 73.3\% and all-head Spearman correlations of 0.677--0.886. The overlap and causal interventions suggest that full-sequence OCR operates as dense sequential multimodal copy-and-paste, repeatedly retrieving visual evidence and routing it to the current output position. Finally, we examine how this shared circuit changes as a general VLM becomes an OCR specialist. Matched base-to-specialized comparisons show that OCR specialization largely preserves head identity, retaining 17--20 of the top 20 heads per task with all-head rank correlations of 0.874--0.942, while redistributing their functional and causal strengths.
\end{abstract}

\section{Introduction}

\begin{wrapfigure}{r}{0.48\textwidth}
\centering
\includegraphics[width=\linewidth]{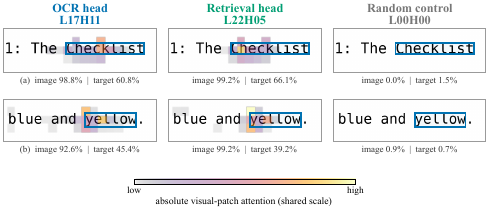}
\caption{Token-level visual attention during Qwen2-VL-2B~\citep{qwen2vl} OCR for generated
tokens (a) ``Checklist'' and (b) ``yellow.'' Columns compare the top text-OCR head, the
retrieval head L22H05, and a head outside both
rankings. Blue boxes mark
target evidence, and all heatmaps use one absolute scale.}
\label{fig:intro_teaser}
\end{wrapfigure}

End-to-end OCR aims to transcribe diverse visible content in an image: characters in running text, cell contents and structure in tables, and glyphs and syntax in formulas~\citep{donut,blecher2023nougatneuralopticalunderstanding,poznanski2025olmocrunlockingtrillionstokens}. Mechanistically, this transcription requires the model to repeatedly locate the visual evidence relevant to the current output and route that evidence into generation. In language models, \emph{retrieval heads}~\citep{wu2024retrieval} perform a closely related operation: they locate answer-bearing positions in context and support copying their contents into the output. The similarity suggests that visual transcription may build on a more general retrieval-and-copy mechanism. This leads to our central question: \emph{does OCR specialization construct a new attention-head circuit, or reconfigure one that is already present in the model?}

Prior work has identified attention heads involved in localized visual text retrieval in general VLMs, using tasks such as passkey retrieval and OCR-VQA~\citep{baek2025ocrheads}. Those experiments reveal question-conditioned visual retrieval, but do not address full-sequence transcription. In our setting, OCR requires repeatedly retrieving the visual evidence needed at each generation step, across running text, visible table content, and mathematical objects.

To study attention-head mechanisms under full transcription, we develop an evidence-grounded protocol that aligns correctly generated tokens with their corresponding visual evidence, with disjoint head discovery and held-out causal validation. Across GLM-OCR~\citep{duan2026glmocr}, MinerU2.5~\citep{niu2026mineru}, and PaddleOCR-VL-1.6~\citep{zhang2026paddleocrvl16}, we identify sparse and reproducible OCR-head sets whose targeted ablation causes substantially greater degradation than matched random ablations across text, tables, and formulas.

Having established sparse and causally relevant OCR heads, we next ask whether they reflect a dedicated reading circuit or reuse a mechanism already present in the model. We therefore compare OCR heads identified from visual transcription with retrieval/copy heads independently identified from text-only retrieval tasks~\citep{wu2024retrieval}. As Figure~\ref{fig:intro_teaser} illustrates, both types of heads attend to the visual evidence corresponding to the current output token, whereas a control head does not. This correspondence extends beyond individual examples, with strong overlap between the two head populations and supporting cross-task causal effects. These results suggest that full-sequence OCR recruits a shared copy-and-route mechanism, operating as dense sequential multimodal retrieval and copy.

If full-sequence OCR largely reuses a pre-existing retrieval-and-copy mechanism, what changes when a general VLM is specialized for OCR? Matched comparisons before and after OCR tuning show a consistent pattern across model generations and OCR tasks: specialization largely preserves the dominant OCR heads and their layer organization, while redistributing their relative functional and causal importance. This suggests that OCR specialization primarily reweights an existing retrieval-and-copy circuit rather than constructing a new attention-head circuit.

Our contributions are threefold:
\begin{enumerate}

\item We identify and causally validate sparse attention-head populations that support full-sequence OCR across text, tables, and formulas in three specialized OCR models.

\item We uncover substantial reuse between OCR heads and independently identified textual retrieval/copy heads in general VLMs, providing mechanistic evidence that full-sequence transcription operates through dense sequential multimodal retrieval and copy.

\item We show that OCR specialization largely preserves the underlying head population and its layer organization, while redistributing the functional and causal importance of the conserved heads.

\end{enumerate}

\section{Related Work}

\paragraph{Document OCR.}
Generative document OCR has evolved from early image-to-sequence models such as Donut~\citep{donut} and Nougat~\citep{blecher2023nougatneuralopticalunderstanding} toward unified page-level transcription of heterogeneous document content. More recent systems such as GOT-OCR2.0~\citep{wei2024generalocrtheoryocr20}, olmOCR~\citep{poznanski2025olmocrunlockingtrillionstokens}, and DeepSeek-OCR~\citep{wei2025deepseekocrcontextsopticalcompression} directly generate page-level text or markup, broadening OCR from local recognition to end-to-end document transcription. In parallel, models such as MinerU2.5~\citep{niu2026mineru} introduce hierarchical or coarse-to-fine processing to handle high-resolution pages efficiently, while GLM-OCR~\citep{duan2026glmocr} and PaddleOCR-VL-1.6~\citep{zhang2026paddleocrvl16} further specialize vision--language architectures for unified document recognition. Across these directions, prior work primarily focuses on recognition quality, structured decoding, resolution handling, and efficiency; we instead study the internal mechanism that supports sustained transcription and how it changes under OCR specialization.

\paragraph{Functional attention heads.}
Prior work has identified sparse attention heads associated with specialized operations, including induction, copying, and long-context retrieval~\citep{voita-etal-2019-analyzing,olsson2022incontextlearninginductionheads,wu2024retrieval}. Recent multimodal studies extend this line to visual grounding and retrieval, finding head-level mechanisms that route textual or visual evidence across modalities~\citep{golovanevsky-etal-2025-vlms,li2026multimodalretrieval,sun2026mechanisticinsightsfunctionalsparsity}. Most directly, \citet{baek2025ocrheads} identify OCR-related attention heads in general VLMs using localized text-reading tasks. We extend this line of work to full-sequence transcription, where visual evidence must be repeatedly routed throughout the generation of text, tables, and formulas. We independently identify OCR and textual retrieval/copy heads, test their functional relationship through causal interventions, and measure how their organization changes under OCR specialization.

\paragraph{Causal validity.}
Attention-based alignment provides a criterion for identifying candidate functional heads, but candidate heads may be behaviorally redundant. Moreover, ablation-based importance can depend on the intervention baseline and replacement value~\citep{li2024optimal}. We therefore use alignment for discovery and assess functional contribution separately through held-out autoregressive ablations against matched random head sets.

\section{Identifying OCR Heads in Full-Sequence Transcription}
\label{sec:ocr-head-discovery}

Let a VLM encode an image as a visual-token prefix and autoregressively generate a token sequence $y_{1:T}$.
For decoder layer $\ell$, attention head $h$, and generated token $y_t$, let
$A_t^{\ell,h}(j)$ denote the attention weight assigned by head $(\ell,h)$ from the query used to generate $y_t$ to the key at position $j$. The key position $j$ ranges over all positions available to that query, including the visual-token prefix and previously generated textual positions.
For each visible reference span $z$, we define its visual evidence set $E(z)$ as the set of visual-token positions whose patch rectangles have IoU greater than $0.1$ with the annotated visual regions corresponding to $z$. These regions include characters in running text, visible cell text in tables, and rendered glyphs or mathematical objects in formulas. Structural markup or layout elements that do not correspond to transcribed textual or glyph content are excluded from the evidence sets. For example, HTML table structure elements, such as tags and border or separator rules, are not treated as visual evidence.

Given these visual evidence sets, we first obtain the model prediction through greedy autoregressive decoding. We then replay the generated sequence to record the attention distributions along the same autoregressive trajectory. During this replay, each position is conditioned on the exact tokens produced by the original decoding process, thereby reconstructing the model's own autoregressive prefixes without substituting reference tokens.
 
We align the decoded prediction with the reference at the character level and identify maximal contiguous spans in which the prediction and reference match exactly. A generated token $y_t$ is eligible for scoring only if its entire decoded character sequence lies within a single maximal exact-match span. For each eligible token, let $z_t$ denote the corresponding aligned visible span in the reference, with visual evidence set $E(z_t)$. Head $(\ell,h)$ records an OCR hit at position $t$ if the key receiving the largest attention weight among all key $\mathcal{K}_t$ accessible to the query belongs to $E(z_t)$. Formally, 
\begin{equation}
 H_t^{\ell,h} = \mathds{1}\left[ \arg\max_{j \in \mathcal{K}_t} A_t^{\ell,h}(j) \in E(z_t) \right]. 
 \label{eq:ocr-hit}
 \end{equation}

\begin{figure}[t]
\centering
\includegraphics[width=0.98\linewidth]{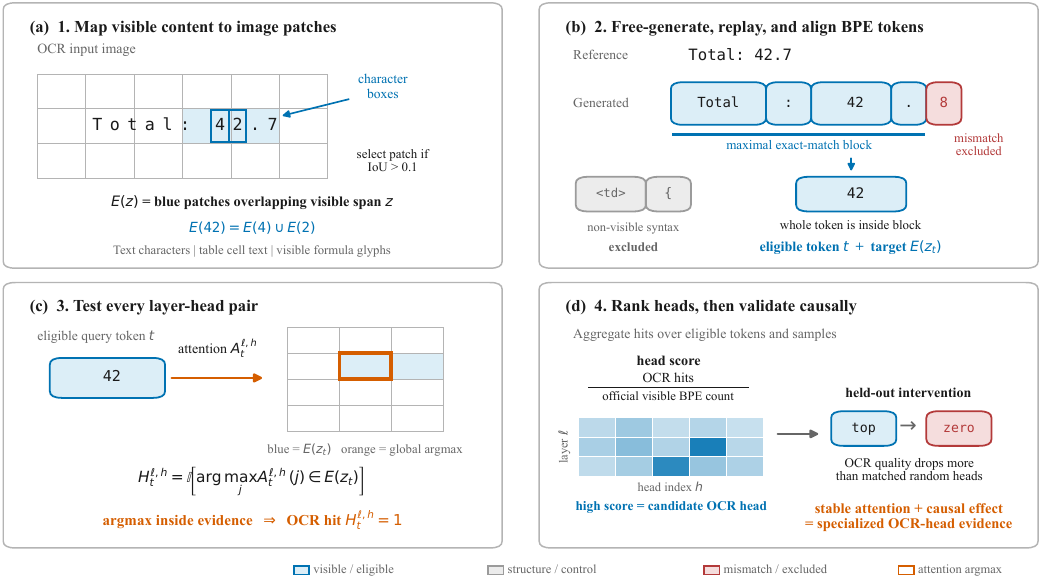}
\caption{Evidence-grounded discovery and causal validation. Visible characters,
cell text, and formula glyphs define evidence patches $E(z)$; only whole BPE
tokens in exact free-generation match blocks are eligible. A head scores a hit
when its global attention argmax falls inside the token evidence. Page-disjoint
discoveries are then tested by held-out intervention against matched random
sets.}
\label{fig:token-evidence-alignment}
\end{figure}

For sample $i$, let $C_i^{\ell,h}$ denote the number of OCR hits recorded by head $(\ell,h)$, and let $G_i$ denote the number of tokens in the official visible-content reference. We define the sample-level OCR score and its average over $N$ samples as
\begin{equation}
 s_i^{\ell,h} = \frac{C_i^{\ell,h}}{\max(G_i,1)}, \qquad \bar{s}^{\ell,h} = \frac{1}{N}\sum_{i=1}^{N}s_i^{\ell,h}. 
 \label{eq:ocr-score}
 \end{equation}
Following prior work~\citep{baek2025ocrheads}, head $(\ell,h)$ is considered active on sample $i$ when $s_i^{\ell,h}>0.1$. We define the activation frequency of head $(\ell,h)$ as
\begin{equation}
 f^{\ell,h} = \frac{1}{N} \sum_{i=1}^{N} \mathds{1}\left[s_i^{\ell,h}>0.1\right]. 
 \label{eq:ocr-activation-frequency}
 \end{equation}
We identify candidate OCR heads as those satisfying both $\bar{s}^{\ell,h}>0.1$ and $f^{\ell,h}\geq 0.1$. These thresholds define the candidate OCR-head set used in threshold-based analyses. For rank-based analyses, we additionally rank all attention heads according to $\bar{s}^{\ell,h}$, irrespective of whether they satisfy the candidate thresholds. Figure~\ref{fig:token-evidence-alignment} summarizes the discovery procedure, from visual-evidence construction and autoregressive replay to token-level OCR hits and head-level scoring.

\section{Validating OCR Heads in Specialized OCR Models}
\subsection{Experimental Setup and Validation Protocol}

We validate our OCR-head identification procedure on GLM-OCR, MinerU2.5, and PaddleOCR-VL-1.6 using a shared OmniDocBench subset of 1,200 instances (400 each for text, table, and formula recognition). Text uses plain transcription, while tables and formulas follow each model’s native output format. For each task, the 400 instances are split into three disjoint discovery folds, A and B (160 each), and a held-out causal-evaluation fold C (80). Heads are discovered independently on A and B, with agreement measured by all-head rank correlation and top-k overlap; consensus heads are evaluated exclusively on C.

Causal damage is measured relative to the intact model using NED increase for text, TEDS~\citep{teds} decrease for tables, and CDM~\citep{cdm} decrease for formulas. For all three tasks, five random head sets exactly match the target set's head count in every layer and exclude the task's consensus top-20 heads. Full evaluation details and controls are provided in Appendix~\ref{app:setup}.

\subsection{Cross-Fold Stability and Held-Out Causal Validation}

We first assess whether OCR-head discovery yields consistent high-ranking
heads across independent data subsets. Across all nine model--task
combinations, the independently discovered top-20 sets overlap by at least
19/20 heads across the two disjoint discovery folds A and B. This strong cross-fold
agreement indicates that the highest-ranked OCR heads are stable across data
subsets rather than driven by a particular fold. Full head-score maps,
per-model overlap statistics, and cross-task intersections are provided in
Appendix~\ref{app:discovery}.

We then ask whether this stable alignment corresponds to a functional
contribution. Following standard head-ablation analyses~\citep{voita-etal-2019-analyzing}, we intervene on the consensus OCR heads and evaluate their effects on the held-out fold $C$, which is excluded from head selection, against matched random head sets.

Targeted ablation of the discovered OCR heads causes consistently and substantially larger performance degradation than layer-matched random interventions across all three specialized OCR models and all three content types. For example, in GLM-OCR, ablating the discovered table heads produces a TEDS drop of $0.903$, compared with $0.032$ for matched random sets, while formula CDM drops by $0.846$, compared with only $0.016$ under random ablation. Similar selectivity is observed in MinerU2.5 and PaddleOCR-VL-1.6, indicating that the effect is not specific to a single OCR architecture. Exact post-ablation scores and additional causal results are provided in Appendix~\ref{app:causal}.

The held-out interventions establish the causal relevance of the discovered OCR-head populations. Since full-sequence transcription repeatedly requires retrieving visual evidence and routing its content to the current output position, we next ask whether these OCR heads reuse the retrieval/copy mechanism already present in general VLMs.

\FloatBarrier
\section{Do OCR Heads Reuse Retrieval/Copy Heads?}
\subsection{Independent OCR and retrieval/copy head discovery}
We use Qwen2-VL-2B~\citep{qwen2vl} for the primary generalist analysis and Qwen3-VL-2B~\citep{qwen3vl} for cross-generation replication. Across both models, we independently identify OCR heads from visual transcription and retrieval/copy heads from text-only needle-in-a-haystack (NIAH) tasks~\citep{wu2024retrieval}.

\begin{figure}[t]
\centering
\includegraphics[width=0.98\linewidth]{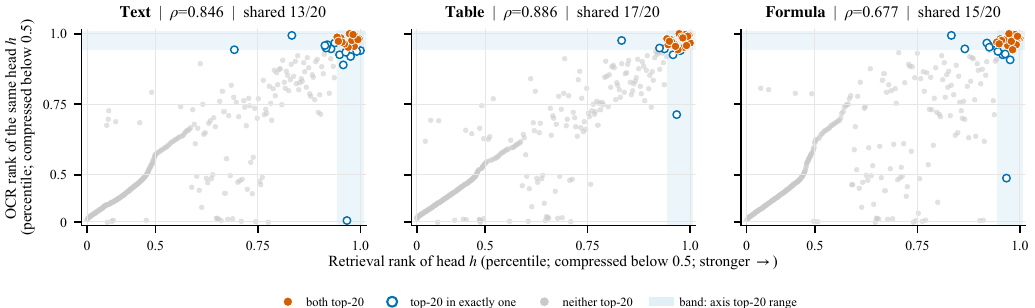}
\caption{Structural correspondence between independently identified OCR and retrieval/copy heads in Qwen2-VL-2B. Each point represents the same head's retrieval and task-specific OCR rank percentiles. Orange points belong to both top-20 sets, blue rings to exactly one, and gray points to neither. Shaded bands mark the top-20 ranges. $\rho$ denotes all-head Spearman correlation; ranks below the 50th percentile are compressed for readability.}
\label{fig:rank-compare}
\end{figure}

The retrieval/copy criterion follows the standard NIAH setup, where a head is considered active when its maximum attention falls within the needle span and the attended source token matches the generated token, thereby operationalizing exact token-level copying from textual context. The OCR criterion follows the same source-selection principle, using spatial alignment to \(E(z_t)\) as the visual analogue of exact source-token matching.

The two head rankings are obtained independently and fixed before any cross-task comparison. We quantify their correspondence using all-head rank correlation, top-$k$ overlap, Jaccard similarity, and asymmetric containment $|O\cap R|/|O|$, where $O$ and $R$ denote the OCR and retrieval/copy head sets, respectively. A separate held-out needle is reserved for causal evaluation.

% \subsection{Structural correspondence between OCR and retrieval/copy heads}
% \label{sec:ocr-retrieval-structural}

% \begin{wrapfigure}{R}{0.43\textwidth}
% \centering
% \includegraphics[width=\linewidth]{figures/ocr_retrieval_correspondence.pdf}
% \caption{Threshold-selected OCR heads contained in independently identified
% textual retrieval/copy sets across two generalist VLMs. Columns denote the
% OCR discovery task.}
% \label{fig:ocr-retrieval-correspondence}
% \end{wrapfigure}

This comparison reveals substantial structural correspondence between the two head populations. In Qwen2-VL, the all-head rankings correlate strongly across text, table, and formula OCR, with Spearman correlations of 0.846, 0.886, and 0.677, respectively.

Figure~\ref{fig:rank-compare} provides a head-level view of this correspondence in Qwen2-VL. Each point represents the same attention head positioned by its retrieval/copy rank and task-specific OCR rank, with the shared top-ranked heads concentrated in the upper-right region. The independently obtained top-20 rankings overlap by 13, 17, and 15 heads for text, tables, and formulas, respectively. This pattern replicates structurally in Qwen3-VL: its 30-head retrieval/copy set yields top-20 overlaps of 15, 15, and 13 heads across the three OCR tasks, with all-head Spearman correlations ranging from 0.806 to 0.859. In addition, 74.1--84.6\% of the selected OCR heads fall within the retrieval/copy set.

All six top-20 overlaps are well above random expectation, confirming that the structural correspondence is consistent across both models and all three OCR tasks.

\subsection{Cross-task causal correspondence}

Structural correspondence provides strong evidence that OCR and retrieval/copy heads may share a common mechanism, but structural overlap alone cannot determine whether they play causally related roles. We therefore examine cross-task causal transfer by intervening on heads identified in one task and evaluating their effects on the other.

In Qwen2-VL, ablating retrieval/copy heads substantially degrades OCR performance. Removing the retrieval top-20 increases text NED from 0.1106 to 0.6234 and reduces formula CDM from 0.9614 to 0.3502, while removing the retrieval top-20 lowers table TEDS from 0.6720 to 0.1597. The symmetric cross-task intervention further supports this functional relationship: ablating the top-20 OCR-head sets identified for text, table, or formula recognition reduces held-out retrieval ROUGE-1 recall from 0.9056 to 0.3556, 0.4000, and 0.2500, respectively.

To approximately match Qwen2-VL's top-20 budget ($20/336\approx6\%$), we additionally ablate 27 of Qwen3-VL's 448 decoder query heads. Although its OCR-ranked top-20 sets leave retrieval recall unchanged, top-27 ablation reveals cross-task effects. Ablating retrieval heads increases text NED from 0.012 to 0.166 and reduces formula CDM from 0.970 to 0.528. Table TEDS also falls from 0.622 to 0.126. Conversely, ablating OCR heads ranked for text, table, and formula recognition reduces retrieval recall from 1.000 to 0.967, 0.406, and 0.339, respectively.

These cross-task interventions support functional reuse of a shared retrieval/copy operation across model generations, while showing that its causal contribution is more concentrated in Qwen2-VL and more budget-sensitive in Qwen3-VL. OCR heads are therefore better viewed as task-active components within a broader retrieval/copy mechanism, with substantial but incomplete correspondence to retrieval/copy heads rather than a separate OCR-specific circuit.

\subsection{Specificity beyond generic visual importance}
The structural and causal correspondence above could still admit a
simpler explanation: the overlapping heads may be generally important
for visual grounding rather than specifically involved in retrieval and
copying. We therefore evaluate them on POPE~\citep{li-etal-2023-pope}, which requires visual grounding but neither
transcription nor source copying. Across all three task-specific overlap
sets, targeted ablation does not significantly exceed layer-matched
random controls, with all confidence intervals for the excess effects
including zero. Thus, generic visual importance alone is unlikely to
explain the OCR--retrieval correspondence (full results in
Appendix~\ref{app:causal}).

\subsection{Counterfactual source-identity transmission}
\label{sec:counterfactual-source-patching}

Structural overlap and cross-task ablation establish that OCR and textual retrieval rely on a shared, causally relevant head population. They do not, however, reveal what these heads transmit or whether their contribution is specific to the visual source associated with the current output token. We therefore use counterfactual Value patching to address two complementary mechanistic questions. First, when the attention pattern is held fixed, do the OCR--retrieval overlap heads transmit information about the identity of the selected visual source? Second, is this transmission specific to that source location, or do the same heads propagate arbitrary visual Value content toward the output?

For each OCR task, we construct minimal image pairs that differ only in the identity of a single target glyph while preserving its spatial location and surrounding content. We retain a pair only when the model correctly recognizes both images, the autoregressive prefixes before the target token are identical, both target identities correspond to a single tokenizer token, the clean next-token prediction is correct on both sides, and the two images share a non-empty visual evidence set for the target position. This yields 184 text, 105 table, and 128 formula pairs. Additional construction details are provided in the appendix~\ref{app:source-patching-case-study}.

Let $A$ denote the recipient image and $B$ the donor image, with target tokens $y_A$ and $y_B$, respectively. For an overlap head $h$ and target generation step $t$, its output can be written as
\begin{equation}
o_t^h = \sum_j \alpha_{tj}^h V_j^h W_O^h ,
\end{equation}
where $j$ indexes the source positions available to the query.
 
\paragraph{Source-identity transmission under fixed attention.}Our first intervention asks whether these heads carry information about
\emph{what} is present at the visual source selected for the current
output token. We keep the recipient attention pattern fixed and replace
only the Value contribution at the target visual evidence positions
$E(z_t)$:
\begin{equation}
\widetilde{o}_t^h = \sum_{j\notin E(z_t)} \alpha_{tj}^{A,h} V_j^{A,h} W_O^h +
\sum_{j\in E(z_t)}
\alpha_{tj}^{A,h} V_j^{B,h} W_O^h .
\label{eq:counterfactual-value-patch}
\end{equation}

Thus, the recipient determines \emph{where} the head attends, while the
donor changes only the Value content contributed by the selected source
region. Queries, keys, attention weights, and all non-target Value
contributions remain those of the recipient. We jointly patch the
task-specific OCR--retrieval overlap heads.

For $A\leftarrow B$, we measure the change in $\operatorname{logit}(y_B)-\operatorname{logit}(y_A)$ and average it with the reverse direction within each pair. Positive values indicate a donor-directed shift under fixed recipient attention. We compare against 20 random head sets matched in size and per-layer allocation.

As shown in Table~\ref{tab:counterfactual-source-patching}, target-source value replacement through the OCR--retrieval overlap heads produces a large donor-directed shift in all three OCR tasks. The mean shifts are 4.476 for text, 2.425 for tables, and 2.498 for formulas, compared with 0.658, 0.812, and 0.168 for exactly layer-matched random head sets. Pair-bootstrap 95\% confidence intervals for the target-source intervention are $[4.251,4.701]$, $[2.076,2.785]$, and $[2.347,2.648]$, respectively. For every task, a one-sided paired sign-flip test comparing the overlap-head intervention with the within-pair mean of the matched-random sets gives $p<10^{-5}$.

The effect is also consistent at the pair level: every retained pair shifts toward the donor identity in both patch directions. With recipient attention fixed, this supports source-identity transmission through the value pathway rather than a change in attention routing.

\paragraph{Spatial selectivity of Value transmission.}
Having established source-identity transmission, we next ask whether the
same heads transmit visual Value content indiscriminately or whether
their causal effect is tied to the visual source associated with the
current output token. We therefore perform a second intervention using
the \emph{same OCR--retrieval overlap heads and the same recipient
attention}, but replace Values at an equal number of non-target visual
positions while leaving the target-region Values untouched. The
non-target positions are chosen to maximize their minimum
sequence-index distance from $E(z_t)$. Head identity and the number of
patched positions are therefore matched across the two interventions,
although attention mass and Value-difference magnitude are not
explicitly matched.

In sharp contrast to target-source replacement, patching unrelated
visual locations produces essentially no donor-directed effect:
$0.004$, $-0.003$, and $-0.003$ for text, tables, and formulas,
respectively, with all corresponding confidence intervals containing
zero. The same heads therefore do not simply propagate arbitrary visual
Value perturbations toward the output. Their causal contribution depends
on Value content originating from the visual source associated with the
current output token.

\begin{table}[t]
\centering
\small
\setlength{\tabcolsep}{4pt}
\begin{tabular}{lrrrr}
\hline
\textbf{Task} &
\textbf{\#Pairs} &
\textbf{Target source} &
\textbf{Matched heads} &
\textbf{Non-target source} \\
\hline
Text    & 184 & 4.476 & 0.658 &  0.004 \\
Table   & 105 & 2.425 & 0.812 & -0.003 \\
Formula & 128 & 2.498 & 0.168 & -0.003 \\
\hline
\end{tabular}
\caption{
Counterfactual value patching in Qwen2-VL-2B. Values are donor-directed
next-token logit shifts. Target uses OCR--retrieval overlap heads at the
target evidence; Random uses matched random heads; Non-target uses the
same overlap heads at unrelated visual positions.
}
\label{tab:counterfactual-source-patching}
\end{table}

The target-source intervention does not flip the pairwise preference
between $y_A$ and $y_B$, nor does it change the full-vocabulary argmax.
We therefore do not view the overlap heads as sufficient on their own
to emit the copied token. Instead, the two interventions show that they
selectively transmit source-identity information from the target region
under fixed attention, while unrelated visual locations contribute
essentially no effect.

\FloatBarrier
\section{How Does OCR Specialization Change the Circuit?}
\subsection{Matched specialization comparison}
% 这里的"mass"用词很奇怪
\begin{wrapfigure}{R}{0.45\textwidth}
\centering
\includegraphics[width=\linewidth]{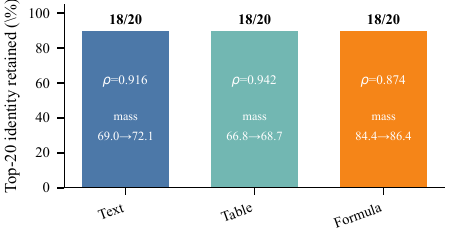}
\caption{Qwen2-VL OCR specialization preserves 18/20 top heads in each task.
Bar height shows identity retention; $\rho$ is the all-head Spearman
correlation, and mass is the pre$\rightarrow$post top-20 OCR-score share.}
\label{fig:specialization-conservation}
\end{wrapfigure}

We next ask how OCR specialization modifies this shared retrieval-and-copy substrate: does specialization introduce a new OCR-specific head organization, or does it primarily reorganize an existing one? To answer this question, we perform matched general-to-specialized comparisons on Qwen2-VL-2B and Qwen3-VL-2B. Within each backbone, we apply the same OCR-head discovery procedure before and after specialization using the same evaluation samples, enabling a direct comparison of head identity and organization.

To distinguish between circuit replacement and circuit reweighting, we examine specialization-induced changes from three complementary perspectives. We first ask whether OCR specialization changes the identity and organization of the dominant heads. We then examine whether specialization alters the distribution of functional importance within this largely preserved head population. Finally, we test whether these organizational changes correspond to shifts in causal contribution through cross-ranked ablations. This decomposition separates changes in which heads are recruited from changes in how strongly they contribute to OCR behavior.

Each configuration uses its own deployment prompt to ensure that the model produces outputs in a fixed and consistent format before and after fine-tuning (see Appendix B for the complete prompts). The matched comparison therefore characterizes the deployed general-to-specialized change, incorporating both parameter adaptation and prompt differences rather than isolating parameter updates alone.

\subsection{OCR-head organization before and after specialization}

We first examine whether OCR specialization changes which heads dominate OCR and how they are organized across layers. In Qwen2-VL, the general and specialized configurations share 18 of their top 20 OCR heads for each of text, table, and formula recognition (Figure~\ref{fig:specialization-conservation} and Table~\ref{tab:prepost-main}). This agreement extends beyond the leading heads: the all-head rankings remain strongly correlated, with Spearman correlations of 0.9164, 0.9423, and 0.8736 for text, tables, and formulas, respectively. The score-weighted layer positions are also stable, shifting by at most 0.34 layers across the three tasks (full statistics in Appendix~\ref{app:prepost}).

The Qwen3-VL comparison exhibits a similar pattern. The general and specialized configurations retain 17/20, 20/20, and 19/20 leading heads for text, tables, and formulas, with corresponding all-head Spearman correlations of 0.9349, 0.9372, and 0.8946. The consistency across two Qwen generations indicates that OCR specialization largely preserves which heads dominate OCR and their coarse organization across layers.

\subsection{OCR-head score redistribution}
Although the dominant OCR-head identities remain largely stable, their relative scores change under specialization. In Qwen2-VL, the effective head count decreases across all three tasks, while the share of the summed OCR scores across all heads accounted for by the top 20 increases from 69.0\% to 72.1\% for text, 66.8\% to 68.7\% for tables, and 84.4\% to 86.4\% for formulas. The decrease in effective head count and the increase in top-20 score mass indicate a more concentrated OCR-score distribution after specialization. Cross-task sharing also increases: the three-way intersection of the text, table, and formula top-20 sets grows from 12 to 14 heads, indicating a larger common core across OCR content types. At the individual-head level, some conserved heads increase in OCR score while others decrease, indicating redistribution rather than uniform amplification.
\subsection{Cross-ranked causal effects}
While OCR-score redistribution reveals how specialization changes the functional emphasis of different heads, we further examine whether these changes correspond to shifts in causal contribution. We perform cross-ranked ablations by independently ranking heads before and after specialization, then ablating both the general-ranked and specialized-ranked head sets in each configuration.  Each ranking is compared with five random head sets exactly matched in head count and per-layer allocation (Table~\ref{tab:app-prepost-causal}).

In Qwen2-VL, both the general- and specialized-ranked top heads remain causally active in both configurations, but the pattern of effects differs across tasks. For text, the specialized-ranked top-20 is more damaging in both configurations, whereas for formulas the general-ranked set remains more damaging. Table effects are closer before specialization and increase for both rankings after specialization. The causal changes therefore do not follow a uniform replacement of one ranking by the other; rather, the relative contribution of the largely conserved head population varies across tasks and configurations.

The Qwen3-VL text results further illustrate how specialization reshapes the causal contributions of existing heads: ablating the general-ranked and specialized-ranked top-20 sets increases NED by $0.3968$ and $0.2904$ before specialization, versus $0.3149$ and $0.4201$ afterward. All four effects exceed their corresponding layer-matched random means. For tables, both sets produce a TEDS drop of $0.2796$ before specialization, versus random means of $0.2272$ and $0.2834$, respectively; afterward, both drops reach $0.4358$, versus $0.0655$ and $0.0391$. For formulas after specialization, the specialized-ranked set causes a larger CDM drop than the general-ranked set ($0.3917$ versus $0.1752$), compared with respective matched random means of $0.0128$ and $0.0705$.

Overall, the cross-ranked ablations suggest circuit conservation with causal reweighting, where OCR specialization largely preserves the underlying head population while altering the distribution of causal contribution within that population.

\section{Conclusion}

Full-sequence OCR relies on sparse, causal head sets, but specialization does
not appear to invent them from scratch. Independently discovered OCR and
retrieval/copy heads overlap across Qwen2-VL and Qwen3-VL, while matched
Qwen2-VL and Qwen3-VL tuning preserves most head identities but reweights their
relative and causal strength. Reading repeatedly recruits a general
copy-and-route substrate; OCR is its dense multimodal use for visual
transcription.

\newpage
\bibliography{iclr2027_conference}
\bibliographystyle{iclr2027_conference}
\appendix
\clearpage
\section{Protocol and Evidence Construction}
\label{app:details}

\paragraph{Visible evidence.}
For text, we re-render official annotations and retain a box for each character;
a BPE token maps to the union of its constituent character evidence rather than
one enclosing word rectangle. For tables, the complete structured output is
preserved, but only visible cell-content tokens enter discovery; HTML/OTSL tags
are excluded. For formulas, rendered glyphs, operators, and composite objects
are aligned to source spans, while braces and non-visible control syntax are
excluded. A patch enters $E(z)$ when its actual rectangle has IoU above 0.1
with at least one visible region for span $z$.

\paragraph{Free-generation replay.}
Greedy generations are replayed using their exact token IDs. We audit replay by
checking whether every replayed position predicts the originally generated
next token. Agreement is 100\% for text and approximately 99.96\% for formulas.
A generated BPE token is scoreable only when its entire decoded span belongs to
a maximal exact reference--prediction match block. This avoids assigning a
visual target to partially correct or mismatched output.

\paragraph{Fold transfer and interventions.}
Heads discovered on A are intervened on B, heads discovered on B on A, and the
A/B consensus on C. Principal free-generation interventions zero head output
only at answer positions. 

\section{Models, Cohorts, and Metrics}
\label{app:setup}

The evaluated models differ in decoder depth and width. GLM-OCR has 16 layers and 16 query heads (256 total); MinerU2.5 has 24 layers and 14 query heads (336); PaddleOCR-VL-1.6 has 18 layers and 16 query heads (288); Qwen2-VL-2B has 28 layers and 12 query heads (336). Each specialist uses 400 instances per task, divided into 160/160/80 by source page. Qwen uses the same sources and splits, with prompt-valid, non-truncated held-out cohorts of 80 text, 75 table, and 79 formula instances.

\paragraph{Prompts and output contracts.}
The OCR specialists are evaluated with their model-native instructions. GLM-OCR
and MinerU2.5 use \texttt{Text Recognition:}, \texttt{Table Recognition:}, and
\texttt{Formula Recognition:}; PaddleOCR-VL-1.6 uses \texttt{OCR:} for text and
the same table and formula instructions. Text targets preserve the visible
reading order, formula targets are delimited \LaTeX{}, and table targets are
canonical HTML. For the native table decoders, the evaluation adapter converts
HTML references to the models' OTSL serialization before token alignment.

The specialized Qwen2-VL and Qwen3-VL checkpoints use the short training-time
prompts \texttt{Text Recognition:}, \texttt{Table Recognition:}, and
\texttt{Formula Recognition:}. For the untuned models, prompt calibration
selects the complete deployment prompts in Table~\ref{tab:qwen2vl-prompts} for
Qwen2-VL and Table~\ref{tab:qwen3vl-prompts} for Qwen3-VL, whose text prompt
is identical to Qwen2-VL's.

\begin{table*}[t]
\centering
\small
\caption{Complete deployment prompts for untuned Qwen2-VL, selected by prompt
calibration.}
\label{tab:qwen2vl-prompts}
\begin{tabular}{@{}lp{0.87\linewidth}@{}}
\toprule
Task & Prompt \\
\midrule
Text &
{\ttfamily Perform OCR on this image. Output only the visible text in natural
reading order, verbatim. Preserve case, punctuation, symbols, spacing, and
line breaks. No commentary, labels, Markdown, or code fences.} \\
\addlinespace
Table &
{\ttfamily Transcribe the table into compact HTML. Return only one
<table>...</table> element with rows in reading order. Use <tr> and <td> or
<th>; preserve every cell's text exactly and use rowspan or colspan only when
visually required. Use no other attributes, prose, Markdown, or code fence.} \\
\addlinespace
Formula &
{\ttfamily Read only the mathematical expression shown. Respond with its exact
LaTeX transcription between \$\$ delimiters. Keep fractions, roots,
superscripts, subscripts, accents, relations, brackets, and punctuation. Add
no explanation.} \\
\bottomrule
\end{tabular}
\end{table*}

\begin{table*}[t]
\centering
\small
\caption{Deployment prompts for untuned Qwen3-VL. The text prompt is identical
to that of Qwen2-VL (Table~\ref{tab:qwen2vl-prompts}).}
\label{tab:qwen3vl-prompts}
\begin{tabular}{@{}lp{0.87\linewidth}@{}}
\toprule
Task & Prompt \\
\midrule
Text &
{\ttfamily Perform OCR on this image. Output only the visible text in natural
reading order, verbatim. Preserve case, punctuation, symbols, spacing, and
line breaks. No commentary, labels, Markdown, or code fences.} \\
\addlinespace
Table &
{\ttfamily Perform table OCR and output only canonical HTML beginning with
<table> and ending with </table>. Preserve row and column order and all
visible cell text. Use only table, tr, td, and th tags plus necessary
rowspan/colspan attributes. Do not add commentary or a code block.} \\
\addlinespace
Formula &
{\ttfamily Transcribe the displayed mathematical formula exactly into LaTeX.
Return one LaTeX string only, enclosed in \$\$...\$\$. Preserve all symbols,
operators, scripts, delimiters, and visible terminal punctuation. Do not
explain and do not use a code fence.} \\
\bottomrule
\end{tabular}
\end{table*}

\paragraph{Decoding and cohort filtering.}
All free-generation runs use greedy decoding (\texttt{do\_sample=False}). The
specialist token budgets are 96, 128, and 2,048 new tokens for text, formula, and
table, respectively; the matched Qwen comparisons use 128, 256, and 2,048.
Discovery retains generations that satisfy the task output contract and do not
reach the token budget. During attention scoring, generated tokens are replayed exactly; only tokens wholly contained in a maximal reference-matching block are eligible, and non-visible syntax is excluded as detailed in Appendix~\ref{app:details}.

For text transcription only, we define
\[
\mathrm{NED}(\hat y,y)=
\frac{\operatorname{Levenshtein}(\hat y,y)}
{\max(|\hat y|,|y|,1)}.
\]
For tables, causal structure damage is baseline TEDS minus intervened TEDS. For
formulas it is baseline CDM minus intervened CDM. 

% 在这里直接新开一段 专门讲specialization的事情
The matched pre/post experiment uses Qwen2-VL-2B which is finetuned by
rank-32 LoRA training (alpha 64) for two epochs on 12,000 mixed OCR examples.
Common valid causal cohorts contain 79 text, 55 table, and 78 formula examples.
All image data, evidence annotations, and generation limits are matched; prompts
follow each configuration's deployment recipe.

The Qwen3-VL-2B replication uses 28 decoder layers and 16 query heads (448 total) with same training configuration. Common valid discovery cohorts contain 319 text, 299 table, and 317 formula examples; held-out causal cohorts contain 79 text, 35 table, and 78 formula examples.  On all 400 examples, base$\to$specialized text NED is 0.0155$\to$0.0168 and table TEDS is 0.7855$\to$0.9192. Thus the checkpoint substantially improves table serialization but does not uniformly improve every task; our mechanism claim rests on head conservation rather than an assumption of monotonic quality gain.

Retrieval/copy discovery uses the same three needles, 20 context lengths, and 10
depths for 600 trials per architecture. Qwen2 and Qwen3 produce 594 and 592
successful trials and identify 31 and 30 retrieval heads, respectively.

\begin{table*}[t]
\centering
\small
\setlength{\tabcolsep}{5pt}
\caption{Retrieval/copy reuse in untuned generalist VLMs. ``OCR in $R$'' is
threshold-set containment; all top-20 overlaps exceed architecture-specific
random expectation with hypergeometric $p<10^{-11}$.}
\label{tab:app-multimodel-retrieval}
\begin{tabular}{llrrrr}
\toprule
Model & OCR task & Retrieval heads & Top-20 overlap & Spearman & OCR in $R$ \\
\midrule
Qwen2-VL & Text & 31 & 13/20 & 0.846 & 92.3\% \\
         & Table & 31 & 17/20 & 0.886 & 93.8\% \\
         & Formula & 31 & 15/20 & 0.677 & 85.7\% \\
\addlinespace
Qwen3-VL & Text & 30 & 15/20 & 0.835 & 74.1\% \\
         & Table & 30 & 15/20 & 0.859 & 84.6\% \\
         & Formula & 30 & 13/20 & 0.806 & 83.3\% \\

\bottomrule
\end{tabular}
\end{table*}
\section{Complete Discovery Results}
\label{app:discovery}
\begin{figure}[t]
\centering
\includegraphics[width=0.92\linewidth]{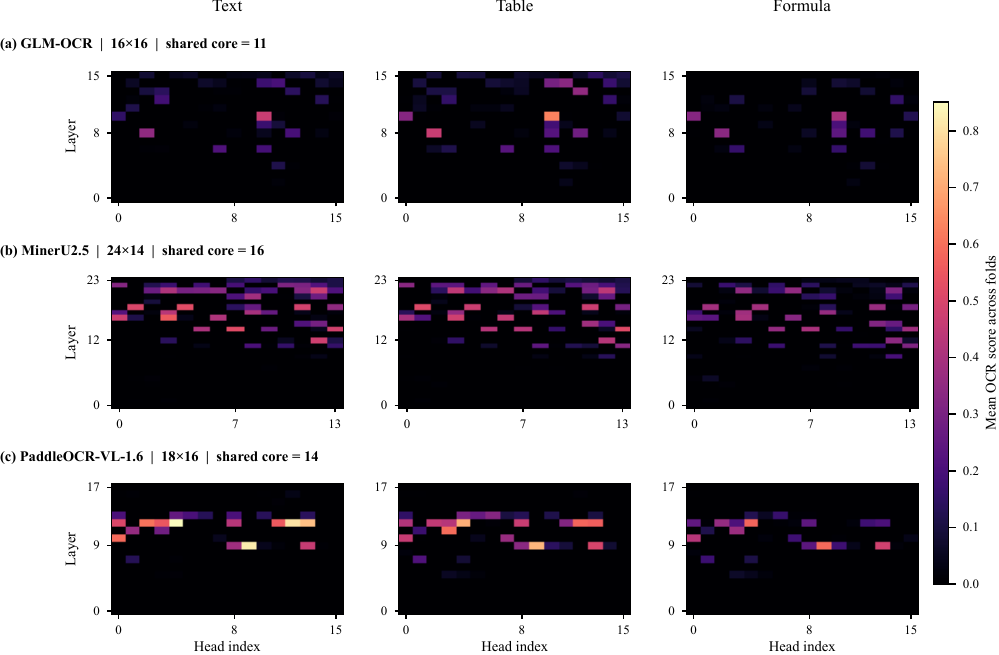}
\caption{Head-wise OCR scores for text, tables, and formulas in three OCR
specialists. High-scoring heads are sparse and architecture dependent, with
task-dependent variation in their locations and strengths.}
\label{fig:discovery-main}
\end{figure}

We further characterize the discovered OCR-head populations through their head-wise score distributions, cross-fold stability, and cross-task overlap. Figure~\ref{fig:discovery-main} shows the full OCR-score maps for text, table, and formula recognition across all three OCR specialists.

\paragraph{Cross-fold stability.}The independently discovered head populations are highly stable across disjoint data folds. For GLM-OCR, the free-generation top-20 sets overlap by 20/20, 20/20, and 19/20 heads for text, tables, and formulas, respectively; MinerU2.5 retains 19/20, 19/20, and 20/20, and PaddleOCR-VL-1.6 retains 19/20, 20/20, and 20/20. Table~\ref{tab:app-stability} additionally reports the teacher-forced diagnostic for GLM-OCR, which yields similarly stable rankings across folds.

\begin{table}[h]
\centering
\small
\caption{GLM-OCR cross-fold discovery stability. TF is the teacher-forced
diagnostic and FG is free-generation replay.}
\label{tab:app-stability}
\begin{tabular}{lccc}
\toprule
Task & TF $\rho$ & TF top-4 & FG top-20 \\
\midrule
Text          & 0.9931 & 4/4 & 20/20 \\
Table content & 0.9951 & 4/4 & 20/20 \\
Formula       & 0.9944 & 4/4 & 19/20 \\
\bottomrule
\end{tabular}
\end{table}

\paragraph{Cross-task sharing.}The task-specific top-20 sets also overlap substantially within each specialist. Their three-way intersections contain 11 heads in GLM-OCR, 16 in MinerU2.5, and 14 in PaddleOCR-VL-1.6, forming the largest Venn partition in all three models (Table~\ref{tab:app-cross-task-overlap}).

\begin{table}[t]
\centering
\footnotesize
\setlength{\tabcolsep}{4pt}
\caption{Exact Venn partitions of the task-specific top-20 OCR-head sets.
Pairwise-only columns contain heads shared by exactly the named two tasks,
excluding the third.}
\label{tab:app-cross-task-overlap}
\begin{tabular}{@{}lccccccc@{}}
\toprule
Model
& \shortstack{Text\\only}
& \shortstack{Table\\only}
& \shortstack{Formula\\only}
& \shortstack{Text--Table\\only}
& \shortstack{Text--Formula\\only}
& \shortstack{Table--Formula\\only}
& \shortstack{All\\three} \\
\midrule
GLM-OCR            & 4 & 2 & 7 & 5 & 0 & 2 & 11 \\
MinerU2.5          & 3 & 1 & 2 & 1 & 0 & 2 & 16 \\
PaddleOCR-VL-1.6   & 3 & 0 & 3 & 3 & 0 & 3 & 14 \\
\bottomrule
\end{tabular}
\end{table}

\section{Additional Causal Results}
\label{app:causal}

\subsection{Full-Sequence OCR Ablation}

We report the exact post-ablation scores for text, table, and formula recognition in the three specialized OCR models in Table~\ref{tab:specialist-causal}. For each model, the targeted top-20 intervention is compared with the intact baseline and the mean over five matched random head sets.

\begin{table}[t]
\centering
\small
\setlength{\tabcolsep}{5pt}
\caption{Held-out causal validation in specialized OCR models.
Top-20 and Random denote the targeted and matched-random ablation conditions,
respectively. Score is the performance after intervention, and $\Delta$ is
the signed change from the intact baseline. Random results are averaged over five head sets exactly matched to the target set's per-layer head counts for all three tasks.}
\label{tab:specialist-causal}
\begin{tabular}{@{}lccccc@{}}
\toprule
& & \multicolumn{2}{c}{Top-20 ablation}
  & \multicolumn{2}{c}{Random ablation} \\
\cmidrule{3-4}
\cmidrule{5-6}
Model & Base & Score & $\Delta$ & Score & $\Delta$ \\
\midrule
\multicolumn{6}{c}{Text (NED)} \\
\midrule
GLM-OCR          & 0.033 & 0.337 & $+0.303$ & 0.086 & $+0.053$ \\
MinerU2.5        & 0.027 & 0.070 & $+0.043$ & 0.026 & $-0.001$ \\
PaddleOCR-VL-1.6 & 0.021 & 0.621 & $+0.600$ & 0.023 & $+0.002$ \\
\midrule
\multicolumn{6}{c}{Table (TEDS)} \\
\midrule
GLM-OCR          & 0.964 & 0.061 & $-0.903$ & 0.932 & $-0.032$ \\
MinerU2.5        & 0.977 & 0.512 & $-0.465$ & 0.862 & $-0.115$ \\
PaddleOCR-VL-1.6 & 0.978 & 0.111 & $-0.867$ & 0.950 & $-0.027$ \\

\midrule
\multicolumn{6}{c}{Formula (CDM)} \\
\midrule
GLM-OCR          & 0.978 & 0.132 & $-0.846$ & 0.962 & $-0.016$ \\
MinerU2.5        & 0.982 & 0.528 & $-0.454$ & 0.983 & $-0.001$ \\
PaddleOCR-VL-1.6 & 0.948 & 0.286 & $-0.662$ & 0.925 & $-0.023$ \\

\bottomrule
\end{tabular}
\end{table}

\subsection{Cross-Model and Cross-Task Effects}

\begin{figure}[t]
\centering
\includegraphics[width=0.98\linewidth]{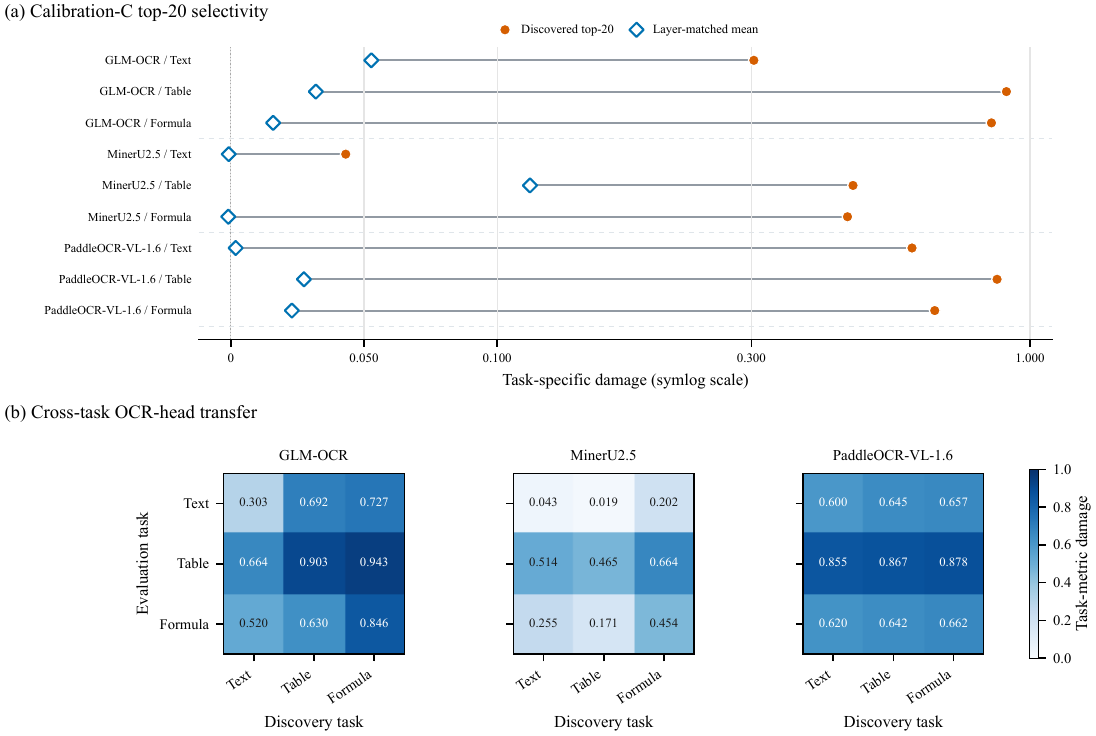}
\caption{Causal effects of OCR-head ablation across three specialized OCR models. (a) Task-specific damage from ablating the discovered top-20 heads and layer-matched random heads. Text, table, and formula use $\Delta$NED, TEDS drop, and CDM drop, respectively. (b) Cross-task transfer of the task-specific top-20 OCR heads. Columns indicate the discovery task and rows the evaluation task.}
\label{fig:app-crossmodel}
\end{figure}

Cross-task effects differ substantially across the three specialists. MinerU is
asymmetric, especially on text targets; GLM-OCR is stronger and more uniform;
and PaddleOCR-VL shows large effects across nearly every cell. This variation
supports a shared routing pool with architecture- and task-dependent strength
rather than one universal profile.

\subsection{Specificity Beyond OCR}
To test whether the overlapping OCR and retrieval heads are broadly important for visual recognition rather than specifically involved in copying, we ablate the three task-specific overlap sets in Qwen2-VL-2B and evaluate object-presence judgments on POPE. Each intervention is compared with 20 random head sets matched in size and per-layer allocation. As shown in Table~\ref{tab:pope_specificity}, targeted ablation reduces accuracy by only 0.22--0.56 percentage points, with no statistically significant excess over the random-control means. These results suggest that the effects observed in OCR and retrieval are not simply a consequence of disrupting generally important visual heads, although they do not establish exclusive involvement in copying.
\begin{table}[t]
\centering
\small
\setlength{\tabcolsep}{3.5pt}
\begin{tabular}{lrrrrr}
\hline
\textbf{Set} & \textbf{\#H} & \textbf{$\Delta$Acc.} &
\textbf{Random} & \textbf{Excess} & \textbf{$p$} \\
\hline
Text--Retrieval    & 12 & 0.56 & 0.16 & 0.39 & .143 \\
Table--Retrieval   & 15 & 0.22 & 0.21 & 0.01 & .667 \\
Formula--Retrieval &  6 & 0.22 & 0.11 & 0.11 & .381 \\
\hline
\end{tabular}
\caption{POPE specificity control. Values are accuracy drops in
percentage points, pooled over Random, Popular, and Adversarial splits.
Random denotes the mean of 20 layer-matched controls; Excess is the
targeted drop minus the random mean.}
\label{tab:pope_specificity}
\end{table}

\section{Token-Level Examples}
\label{app:token}
\begin{figure}[t]
\centering
\includegraphics[width=\linewidth]{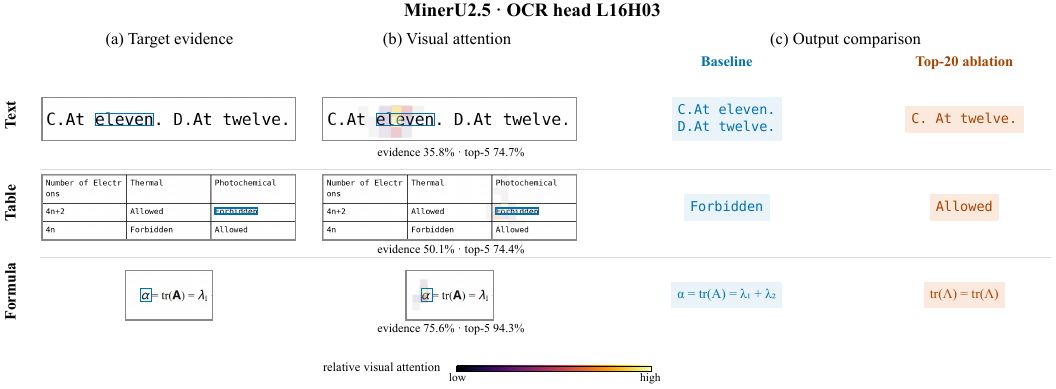}
\caption{Token-level examples in MinerU2.5. Rows show text, table, and formula recognition. (a) Input crops with target evidence outlined in blue. (b) Visual attention of L16H03 at the selected generation step. Evidence and top-5 percentages are normalized by total attention to visual patches; argmax denotes the most-attended visual patch. (c) Corresponding output excerpts before and after task-specific top-20 ablation, with single-head effects reported separately. Input crops show local evidence and need not contain the full output excerpt.}
\label{fig:app-token-examples}
\end{figure}

To complement the aggregate ablation results in specialized OCR models, we examine where an identified OCR head attends during individual generation steps and how the corresponding outputs change under ablation. We use MinerU2.5 head L16H03 (layer index 16, head index 3) as a case study across text, table, and formula recognition. Figure~\ref{fig:app-token-examples} pairs the visual evidence for each selected output token with this head's attention over image patches and compares baseline generation with ablation of either L16H03 alone or the task-specific top-20 OCR heads.

L16H03 assigns 35.8\%, 50.1\%, and 75.6\% of its visual attention to the evidence patches for \texttt{eleven}, \texttt{Forbidden}, and $\alpha$, respectively. Its five most-attended visual patches account for 74.7\%, 74.4\%, and 94.3\% of visual attention, indicating spatial concentration rather than necessarily target alignment. Ablating L16H03 alone produces only spacing or notation changes, with no output change in the table example. In contrast, top-20 ablation replaces \texttt{eleven} with \texttt{twelve} and omits the following option, changes the target table cell from \texttt{Forbidden} to \texttt{Allowed}, and rewrites the formula. These examples distinguish visual alignment by an individual head from the causal contribution of a jointly ablated head set; the content errors cannot be attributed to L16H03 alone.

\section{Matched Pre/Post Specialization Results}
\label{app:prepost}

\begin{table*}[t]
\centering
\small
\setlength{\tabcolsep}{5pt}
\caption{Matched specialization replicates across Qwen generations. Discovery
is computed independently in each configuration on common valid samples.}
\label{tab:app-qwen-generations}
\begin{tabular}{llrrrr}
\toprule
Backbone & Task & Common $N$ & Qualified B$\to$S & Top-20 overlap & Spearman \\
\midrule
Qwen2-VL & Text & 319 & 17$\to$20 & 18/20 & 0.9164 \\
         & Table & 290 & 16$\to$20 & 18/20 & 0.9423 \\
         & Formula & 312 & 8$\to$10 & 18/20 & 0.8736 \\
\addlinespace
Qwen3-VL & Text & 319 & 27$\to$33 & 17/20 & 0.9349 \\
         & Table & 299 & 26$\to$34 & 20/20 & 0.9372 \\
         & Formula & 317 & 18$\to$15 & 19/20 & 0.8946 \\
\bottomrule
\end{tabular}
\end{table*}

\begin{table*}[t]
\centering
\small
\setlength{\tabcolsep}{4pt}
\caption{Held-out cross-ranked causal effects. Cells report performance after
ablation, with signed change from base in parentheses. Text uses NED
($\downarrow$), tables use TEDS ($\uparrow$), and formulas use CDM
($\uparrow$). B/S heads are ranked in the base/specialized configuration.
Rnd mean averages ten layer-matched random control sets (five matched to each
ranking in head count and per-layer allocation).}
\label{tab:app-prepost-causal}
\begin{tabular}{lllrrrr}
\toprule
Model/task & Eval. & $k$ & Base & B heads & S heads & Rnd mean \\
\midrule
Qwen3-VL Text & Base & 5  & 0.012 & 0.022 ($+0.010$) & 0.019 ($+0.006$) & 0.018 ($+0.005$) \\
     &      & 10 & 0.012 & 0.028 ($+0.015$) & 0.018 ($+0.005$) & 0.019 ($+0.006$) \\
     &      & 20 & 0.012 & 0.409 ($+0.397$) & 0.303 ($+0.290$) & 0.020 ($+0.007$) \\
Qwen3-VL Text & Specialized & 5  & 0.019 & 0.022 ($+0.002$) & 0.024 ($+0.005$) & 0.018 ($-0.001$) \\
     &             & 10 & 0.019 & 0.022 ($+0.003$) & 0.022 ($+0.002$) & 0.019 ($0.000$) \\
     &             & 20 & 0.019 & 0.334 ($+0.315$) & 0.439 ($+0.420$) & 0.020 ($+0.001$) \\
\addlinespace
Qwen3-VL Table & Base & 5  & 0.773 & 0.645 ($-0.128$) & 0.666 ($-0.107$) & 0.603 ($-0.170$) \\
      &      & 10 & 0.773 & 0.673 ($-0.100$) & 0.673 ($-0.100$) & 0.579 ($-0.194$) \\
      &      & 20 & 0.773 & 0.493 ($-0.280$) & 0.493 ($-0.280$) & 0.518 ($-0.255$) \\
Qwen3-VL Table & Specialized & 5  & 0.926 & 0.918 ($-0.009$) & 0.931 ($+0.005$) & 0.899 ($-0.027$) \\
      &             & 10 & 0.926 & 0.914 ($-0.012$) & 0.914 ($-0.012$) & 0.888 ($-0.039$) \\
      &             & 20 & 0.926 & 0.491 ($-0.436$) & 0.491 ($-0.436$) & 0.874 ($-0.052$) \\
\addlinespace
Qwen3-VL Formula & Base & 5  & 0.969 & 0.967 ($-0.002$) & 0.968 ($-0.001$) & 0.950 ($-0.019$) \\
        &      & 10 & 0.969 & 0.963 ($-0.007$) & 0.940 ($-0.029$) & 0.956 ($-0.013$) \\
        &      & 20 & 0.969 & 0.880 ($-0.089$) & 0.590 ($-0.379$) & 0.933 ($-0.036$) \\
Qwen3-VL Formula & Specialized & 5  & 0.976 & 0.973 ($-0.003$) & 0.971 ($-0.005$) & 0.967 ($-0.010$) \\
        &             & 10 & 0.976 & 0.975 ($-0.002$) & 0.941 ($-0.035$) & 0.961 ($-0.015$) \\
        &             & 20 & 0.976 & 0.801 ($-0.175$) & 0.584 ($-0.392$) & 0.934 ($-0.042$) \\
\addlinespace
Qwen2-VL Text & Base & 5  & 0.080 & 0.140 ($+0.060$) & 0.118 ($+0.038$) & 0.092 ($+0.012$) \\
     &      & 10 & 0.080 & 0.153 ($+0.073$) & 0.229 ($+0.149$) & 0.099 ($+0.019$) \\
     &      & 20 & 0.080 & 0.415 ($+0.335$) & 0.632 ($+0.552$) & 0.149 ($+0.068$) \\
Qwen2-VL Text & Specialized & 5  & 0.013 & 0.013 ($+0.000$) & 0.012 ($-0.001$) & 0.014 ($+0.002$) \\
     &             & 10 & 0.013 & 0.033 ($+0.020$) & 0.103 ($+0.090$) & 0.021 ($+0.008$) \\
     &             & 20 & 0.013 & 0.273 ($+0.260$) & 0.508 ($+0.495$) & 0.022 ($+0.009$) \\
\addlinespace
Qwen2-VL Table & Base & 5  & 0.613 & 0.428 ($-0.184$) & 0.462 ($-0.151$) & 0.605 ($-0.007$) \\
      &      & 10 & 0.613 & 0.403 ($-0.209$) & 0.298 ($-0.315$) & 0.580 ($-0.032$) \\
      &      & 20 & 0.613 & 0.152 ($-0.460$) & 0.161 ($-0.452$) & 0.392 ($-0.220$) \\
Qwen2-VL Table & Specialized & 5  & 0.842 & 0.579 ($-0.263$) & 0.604 ($-0.238$) & 0.819 ($-0.023$) \\
      &             & 10 & 0.842 & 0.535 ($-0.307$) & 0.368 ($-0.474$) & 0.788 ($-0.054$) \\
      &             & 20 & 0.842 & 0.175 ($-0.667$) & 0.146 ($-0.696$) & 0.576 ($-0.266$) \\
\addlinespace
Qwen2-VL Formula & Base & 5  & 0.959 & 0.705 ($-0.254$) & 0.678 ($-0.281$) & 0.958 ($-0.001$) \\
        &      & 10 & 0.959 & 0.629 ($-0.330$) & 0.477 ($-0.482$) & 0.957 ($-0.002$) \\
        &      & 20 & 0.959 & 0.334 ($-0.625$) & 0.497 ($-0.462$) & 0.952 ($-0.007$) \\
Qwen2-VL Formula & Specialized & 5  & 0.951 & 0.709 ($-0.242$) & 0.683 ($-0.268$) & 0.945 ($-0.006$) \\
        &             & 10 & 0.951 & 0.610 ($-0.340$) & 0.431 ($-0.520$) & 0.943 ($-0.008$) \\
        &             & 20 & 0.951 & 0.304 ($-0.647$) & 0.402 ($-0.549$) & 0.934 ($-0.016$) \\
\bottomrule
\end{tabular}
\end{table*}

The three-task intersection grows from 12 to 14 heads, heads shared by exactly
two tasks fall from 9 to 5, and the task-specific periphery changes from 6 to 8.
L19H06 and L17H01 strengthen across all tasks, whereas L17H08 weakens across all
three. These opposing changes rule out uniform amplification and instead
indicate redistribution within a conserved circuit.

\begin{table*}[t]
\centering
\small
\caption{Matched OCR-head discovery before (G) and after specialization in
Qwen2-VL. Top-head overlap, all-head rank correlation, and score-weighted layer
position show little change in head organization.}
\label{tab:prepost-main}
\begin{tabular}{lrrrrr}
\toprule
Task & Common $N$ & Qualified G$\to$S & Top-20 overlap & Spearman
& Weighted layer G$\to$S \\
\midrule
Text & 319 & 17$\to$20 & 18/20 & 0.9164 & 18.695$\to$18.681 \\
Table & 290 & 16$\to$20 & 18/20 & 0.9423 & 18.754$\to$18.697 \\
Formula & 312 & 8$\to$10 & 18/20 & 0.8736 & 17.227$\to$17.566 \\
\bottomrule
\end{tabular}
\end{table*}

\FloatBarrier
\section{Worked Examples of Source-Identity Replacement}
\label{app:source-patching-case-study}

\begin{figure}[htbp]
\centering
\includegraphics[width=\linewidth]{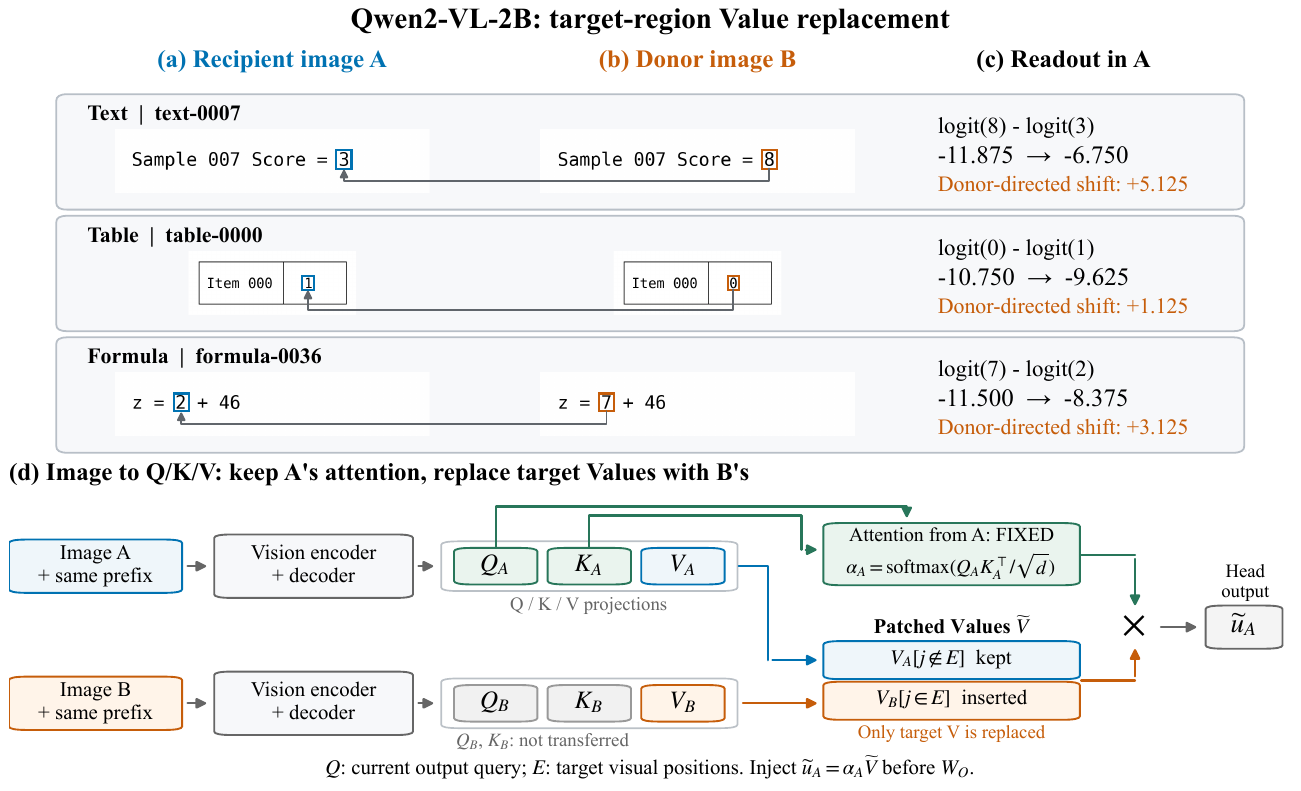}
\caption{Worked examples of $A\leftarrow B$ source-identity replacement. Image boxes mark changed glyphs, not attention intensity; links associate donor and recipient target regions. Panel (d) traces image and prefix processing to Q/K/V: A's attention is fixed (green), non-target Values are retained (blue), and B supplies target Values only (orange). The overlap sets contain 13, 17, and 15 heads for text, tables, and formulas. Logits shift toward the donor without changing the next-token argmax.}
\label{fig:source-patching-case-study}
\end{figure}

Figure~\ref{fig:source-patching-case-study} traces the Value replacement in Section~\ref{sec:counterfactual-source-patching} using controlled Qwen2-VL-2B pairs. Each example is nearest its task's median bidirectional target effect, with ties broken by sample ID.

\paragraph{1. Obtain Q/K/V from paired inputs.}
Image $A$ reads \texttt{Sample 007 Score = 3}; $B$ substitutes \texttt{8} for \texttt{3}. Both are recognized correctly. Processing each image and the same pre-digit output prefix through the vision encoder and decoder yields the current-output query $Q$ and context keys $K$ and Values $V$. We cache A's attention row $\alpha_A$, both runs' target-region Values, and A's clean head output $u_A$. The glyph boxes identify matched visual positions $E$; table and formula pairs follow the same one-digit construction.

\paragraph{2. Fix A's attention and replace only target Values.}
For each overlap head, we construct $\widetilde{u}_A=u_A+\sum_{j\in E}\alpha_{Aj}(V_{Bj}-V_{Aj})$. The cached attention weights determined by A's Q/K remain unchanged: B supplies only target-region Values, not its Q/K. Contributions outside $E$ also remain from A. We rerun A with the same prefix and inject $\widetilde{u}_A$ at the target step, before output projection $W_O$. This fixes attention in the constructed head outputs, not throughout the network: other heads are not overwritten, and downstream computation proceeds normally. Neither pixels nor output tokens are replaced.

\paragraph{3. Read the donor-directed logit shift.}
After injection, the text contrast $\operatorname{logit}(8)-\operatorname{logit}(3)$ changes from $-11.875$ to $-6.750$: a $+5.125$ shift, versus $+0.438$ averaged over 20 layer-matched controls. Table and formula shifts are $+1.125$ and $+3.125$, versus control means of $+0.188$ and $+0.163$. Wrong-location shifts are zero in all three cases. These are $A\leftarrow B$ results; the main analysis averages both directions. The shifts support donor-content transmission under fixed attention, but do not change the predicted token in these examples.

\end{document}

%% file: math_commands.tex
\usepackage{amsmath,amsfonts,bm}

\def\eqref#1{equation~\ref{#1}}
\def\1{\bm{1}}

\DeclareMathAlphabet{\mathsfit}{\encodingdefault}{\sfdefault}{m}{sl}
\SetMathAlphabet{\mathsfit}{bold}{\encodingdefault}{\sfdefault}{bx}{n}